\documentclass[numbered]{trbunofficial}
\makeatletter
\let\quickwordcount\@gobble        

\providecommand{\TotalWords}[1]{}  
\TotalWords{0}                     

\ifcsname c@wordcounter\endcsname\else
  \newcounter{wordcounter}
\fi
\makeatother

\makeatletter
\renewpagestyle{main}{%
  \sethead{\@AuthorHeaders}{}{\thepage}
  \setfoot{}{\scriptsize Transportation Research Board 104th Annual Meeting, Washington, D.C.}{}
}
\makeatother

\usepackage{graphicx}
\usepackage{newtxtext,newtxmath} 
\usepackage{float}
\usepackage{placeins}
\usepackage{tabularx}
\usepackage{color}
\usepackage{booktabs}

\usepackage[hidelinks]{hyperref}

\AuthorHeaders{Pawar, Prozzi, Hong, and Congress}
\title{Deep Learning Framework for Infrastructure Maintenance: Crack Detection and High-Resolution Imaging of Infrastructure Surfaces}

\author{%
  \textbf{Nikhil M. Pawar}\\
  Maseeh Department of Civil, Architectural and Environmental Engineering\\
  The University of Texas at Austin, Austin, Texas (US), 78712 \\
  nikhil.pawar@austin.utexas.edu \\
  ORCID: \url{https://orcid.org/0000-0002-1161-3289} \\[1em]
  \textbf{Jorge A. Prozzi}\\
  Maseeh Department of Civil, Architectural and Environmental Engineering\\
  The University of Texas at Austin, Austin, Texas (US), 78712 \\
  prozzi@mail.utexas.edu \\
  ORCID: \url{https://orcid.org/0000-0003-0489-5879} \\[1em]
  \textbf{Feng Hong}\\
  Ingram School of Engineering\\
  Texas State University, San Marcos (US), 78666 \\
  fenghong@txstate.edu \\
  ORCID: \url{https://orcid.org/0000-0002-2062-4772} \\[1em]
  \textbf{Surya Sarat Chandra Congress}\\
  Department of Civil and Environmental Engineering\\
  Michigan State University, East Lansing (US), 48824 \\
  surya@msu.edu \\
  ORCID: \url{https://orcid.org/0000-0001-5921-9582}
}

\begin{document}
\maketitle

\section{Abstract}
Recently, there has been an impetus for the application of cutting-edge data collection platforms such as drones mounted with camera sensors for infrastructure asset management. However, the sensor characteristics, proximity to the structure, hard-to-reach access, and environmental conditions often limit the resolution of the datasets. A few studies used super-resolution techniques to address the problem of low-resolution images. Nevertheless, these techniques were observed to increase computational cost and false alarms of distress detection due to the consideration of all the infrastructure images i.e., positive and negative distress classes. In order to address the pre-processing of false alarm and achieve efficient super-resolution, this study developed a framework consisting of convolutional neural network (CNN) and efficient sub-pixel convolutional neural network (ESPCNN). CNN accurately classified both the classes. ESPCNN, which is the lightweight super-resolution technique, generated high-resolution infrastructure image of positive distress obtained from CNN. The ESPCNN outperformed bicubic interpolation in all the evaluation metrics for super-resolution. Based on the performance metrics, the combination of CNN and ESPCNN was observed to be effective in preprocessing the infrastructure images with negative distress, reducing the computational cost and false alarms in the next step of super-resolution. The visual inspection showed that EPSCNN is able to capture crack propagation, complex geometry of even minor cracks. The proposed framework is expected to help the highway agencies in accurately performing distress detection and assist in efficient asset management practices.
\hfill\break%
\noindent\textit{Keywords}: Convolutional Neural Network, Efficient Sub-pixel Convolutional Neural Network, Super-resolution, Crack Detection
\newpage

\section{Introduction}
According to  2021 Infrastructure Report Card by \citet{ASCE2021}, the infrastructure score received was C- which is improvement in last 20 years. However, still 40\% of roads are in poor or mediocre condition, where the number remains stagnant over past several years \cite{ASCE2021}. This highlights the importance of identifying distresses and performing timely maintenance, which are crucial for the resilience and safety of civil infrastructure. Traditionally, distresses are identified through visual manual inspection by an infrastructure inspector, which can be laborious and time-consuming \cite{spencer2019advances,ali2022structural}. Moreover, there is a risk of human error in missing difficult-to-spot distress, which can worsen over time and cause further damage to the structure. This can lead to significant safety issues as the integrity of the structure is compromised. For example, the detection of distress in pavement surfaces was traditionally done through manual visual inspection \cite{ragnoli2018pavement,zakeri2017image}, often prone to overlooking difficult-to-spot distresses. This oversight can result in delayed maintenance and rehabilitation action and reduce the serviceability of roads and highways. Other more accurate techniques include the addition of sensor monitoring to visual inspection. The sensors can provide accurate infrastructure information, however, they cannot be expanded to the network level due to setup issues (defective sensors, wiring problems), and battery issues which hinder continuous monitoring \cite{mishra2022structural}. 

With the advent of progress in computer vision, automatic distress detection has become a promising field compared to the traditional manual visual inspection, and sensor data collection \cite{spencer2019advances,ali2022structural}. This method includes collecting the images (or videos) which captures visual information similar to visual inspector, and encodes entire view information which address the challenge of the sensors \cite{spencer2019advances}. In computer vision one of the most fundamental algorithm utilized for distress detection is convolutional neural network (CNN) \cite{ali2022structural,du2021application}.  For example 
\citet{zhang2017automated} proposed a CrackNet which was advanced variant of CNN to detect cracks on 3D asphalt surfaces. The study utilizes 1800 images for training and 200 images for testing. The testing results showed precision of 90.13\%, recall of 87.63\% and F1 score of 88.86\% \cite{zhang2017automated}. The CrackNET outperformed 3D modeling and Pixel-SVM \cite{zhang2017automated}. \citet{ozgenel2018performance} utilized pretrained CNN for crack detection on Concrete Crack Image for Classification (CCIC) dataset. The study was conducted to understand the size of the training dataset, number of epochs for training and extrapolation to other material types for seven pretrained networks \cite{ozgenel2018performance}.   

Other studies include, \citet{yang2020deep}  utilized transfer learning deep convolutional neural network model for crack detection.  The dataset used were CCIC \cite{ozgenel2018performance} , SDNET (containing bridge decks, walls and pavements) \cite{maguire2018sdnet2018} and bridge crack dataset \cite{xu2019automatic}.The results showed that compared to full learning the transfer learning the training time can be reduced 10 time \cite{yang2020deep}. Further, transfer learning can improve the performance by 2.33\% and 5.06\% for SDNET and BCD compared to full learning \cite{yang2020deep}. Some of the recent advance include, \citet{alexander2022fusion} proposed a thermal and RGB images fusion ResNET-based semantic segmentation for damage detection. The findings showed that for crack detection the model trained with fusion of RGB and thermal images increased the IOU by 14\% compared to RGB model alone \cite{alexander2022fusion}. 

However, these studies assume that the collected images are of high resolution, captured by unmanned aerial vehicles (UAVs), robots, and cameras mounted on vehicles such as trucks. Despite this assumption, UAVs often face issues such as vibration and the inability to get close to surfaces due to safety restrictions, which can cause motion blur and result in low-resolution images of infrastructure \cite{ellenberg2016bridge,xiang2022crack}. Furthermore, when the images are taken with low-resolution cameras, such as mobile cameras, accurate distress detection becomes problematic. To address these challenges, super-resolution algorithms \cite{pawar2024esm,pawar2022spatiotemporal,pawar2022complex} can be employed to enhance the resolution of the captured images, thereby facilitating precise distress detection. For instance, \citet{bae2021deep} proposed a two-phase crack detection known as SrcNet algorithm. The SrcNet consists of super-resolution of digital images from bridge surfaces acquired through a climbing robot and UAV in phase 1, and includes an encoder decoder network for crack detection of super-resolution images in phase 2 \cite{bae2021deep}. The findings showed that compared to raw digital image models, SrcNet had 24\% higher performance \cite{bae2021deep}. Another study used SrcNet based bridge crack evaluation via hybrid image matching \cite{jang2022automated}. The matching involves the fusion of laser-induced infrared thermography images and vision images through super-resolution using SrcNet \cite{jang2022automated}. The results showed that false alarms due to harsh bridge conditions reduced up to 49.91 and 13.31\%  in terms  of precision and recall for validation dataset \cite{jang2022automated}. Another study includes a method by proposed by \citet{xiang2022crack} consisting of super-resolution, semantic segmentation  and improved medial axis transform (to quantify length and width of crack) for automatic microcrack detection. For super-resolution among different methods tested, super-resolution feedback network (SRFBN) had the highest peak-signal-to-noise ratio and structural similarity index while having the lowest number of parameters (3,631,478) \cite{xiang2022crack}. For segmentation, CDU-Net had higher accuracy such as F1 score of 84.86\% compared to other networks \cite{xiang2022crack}. 

In all the studies discussed, there is no preprocessing step to exclude images without cracks (identifying negative cases), which increases the computational cost for super-resolution, and false alarms for distress detection. Furthermore, in the comparative study for super-resolution, the most efficient network for super-resolution i.e., (SRFBN) had  large network with 3,631,468 parameters to train \cite{xiang2022crack}. To address the challenge of pre-processing and efficient super-resolution, we propose a framework consisting of a Convolutional Neural Network (CNN) and an Efficient Sub-Pixel Convolutional Neural Network (ESPCNN) . The CNN for the classification of low-resolution infrastructure surfaces and ESPCNN for super-resolution of positive-crack surfaces. CNN is chosen for crack detection because it is a fundamental technique in computer vision and is computationally lightweight compared to transformers and encoders \cite{chen2021review}. The ESPCNN is selected for super-resolution due to its competitive accuracy and computational efficiency, as demonstrated in previous comparative studies of super-resolution techniques \cite{pawar2024esm,soltanmohammadi2023comparative}. Additionally, the ESPCNN has just 83,376 parameters to train, making it much more computationally efficient than SRFBN.

\section{Data Description}
The dataset used in the study is taken from \citet{ozgenel2018performance}, consisting of 40,000 images with two classes, negative and positive distress within concrete structures such as walls, floors, and pavements. In each class, the RGB images have \(227\times227\) pixels. The authors generated a dataset from 458 images with a pixel size of \(4032\times3024\) using the method proposed by \citet{zhang2016road}. For our study, we considered a total of 10,000 RGB images equally distributed between each class. 

For stage 1, i.e., crack classification, the images are resized to \(32\times32\) for each class and are classified as low-resolution (LR) images. These images are divided into training, validation and testing. The training, validation and testing dataset consist of 4900 (49\%), 2100 (21\%), and 3000 (30\%) images, respectively, belonging to both classes. For stage 2 (super-resolution), we consider positive images \(32\times32\)  as low-resolution, and resize the original positive images into \(128\times128\) for high-resolution, while maintaining their aspect ratio for computational efficiency. The training, validation, and test data for ESPCNN consist of 3200, 800, and 1000 images each of low- and high-resolution.

\section{Methodology}
\begin{figure}[H]
    \centering
    \includegraphics[width=0.4\linewidth]{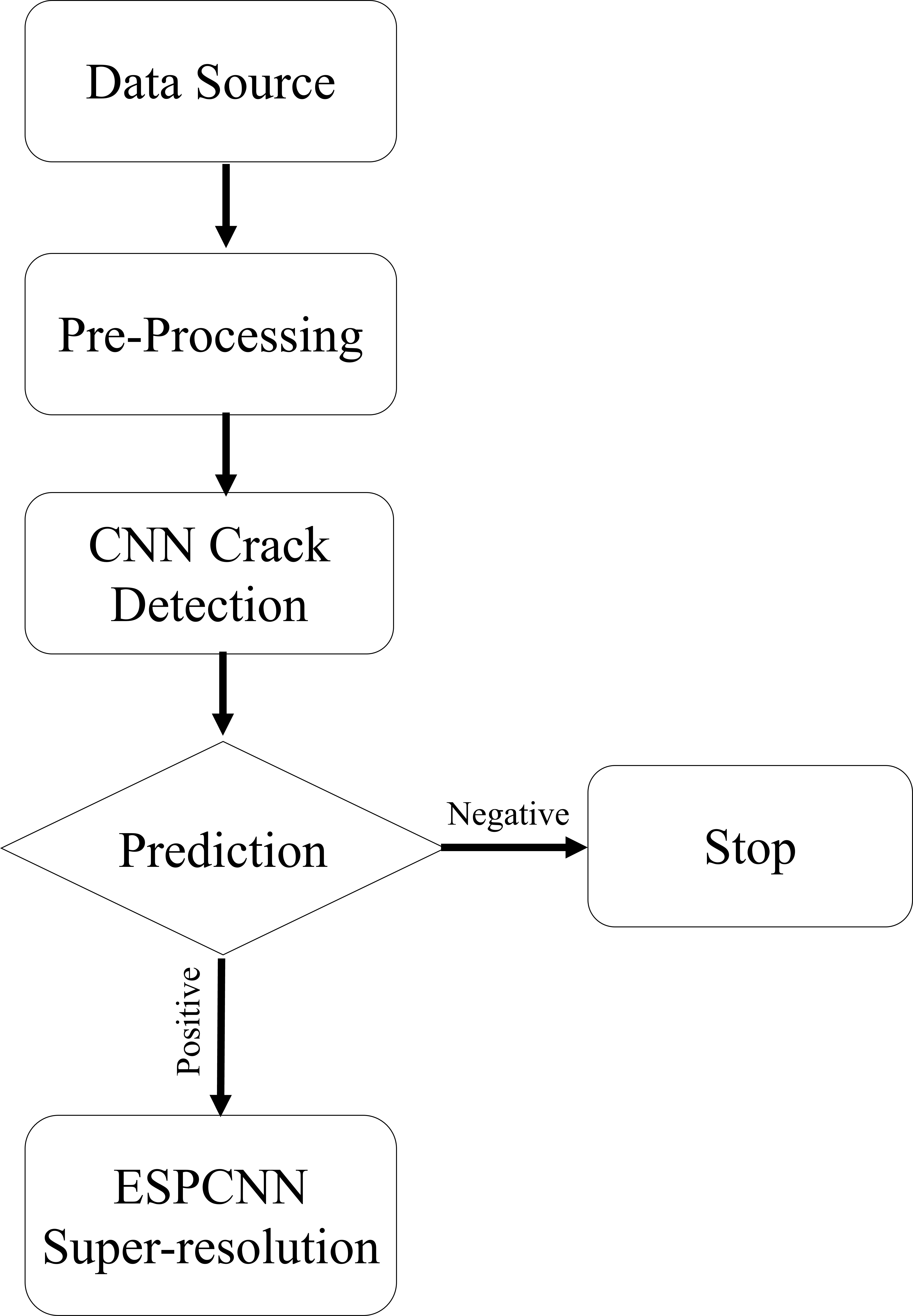}
    \caption{Proposed  framework to address dual challenge of crack detection and super-resolution of infrastructure images.}
    \label{fig:1}
\end{figure}

We propose a deep learning framework that incorporates a convolutional neural network (CNN) and an efficient sub-pixel convolutional neural network (ESPCNN) for crack classification and super-resolution, as depicted in Fig.~\ref{fig:1}. In our framework, the low-resolution infrastructure data is first acquired, pre-processed, and prepared for the subsequent steps. In the next step the low-resolution images are classified as positive and negative crack. If the images are positive, then these positive images are fed into ESPCNN to enhance the resolution of the infrastructure image, thus facilitating better identification of distress.

\FloatBarrier

\subsection{Convolutional Neural Network}
\label{CNN}
\begin{figure}[H]
    \centering
    \includegraphics[width=0.9\linewidth]{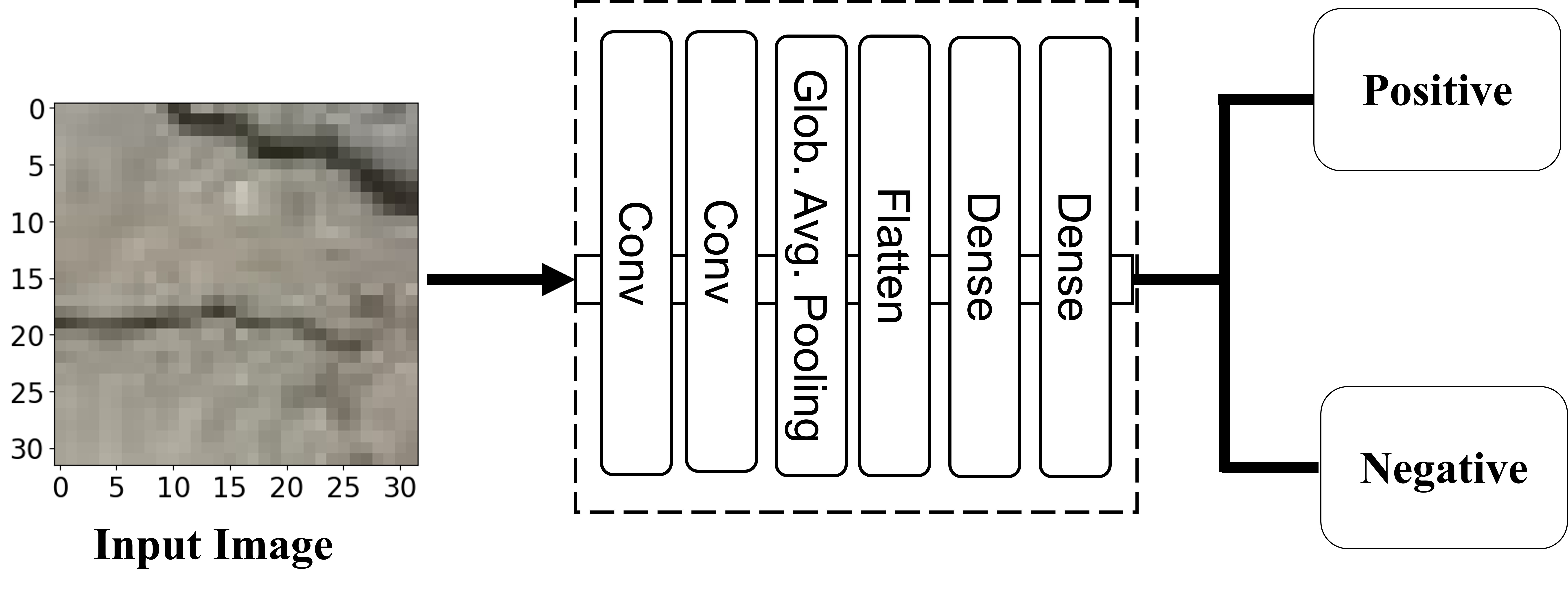}
    \caption{A schematic architecture of a convolutional neural network. The architecture takes the surface image as input and classifies it as either positive (indicating distress) or negative (indicating no distress).}
    \label{fig:2}
\end{figure}

CNN \cite{li2021survey} is most commonly used algorithm used for image classification. A CNN  as shown in Fig.~\ref{fig:2} consists of two convolutional layers, global average pooling layer, flatten and two dense layers. The first convolutional layer consists of 16 filters and kernel size of \(3\times3\), and ReLU activation function. The second layer consists of convolutional layer with 32 filters and kernel size of \(3\times3\),  and ReLU activation function. The third layer, Global average pooling layer, computes the average value of each feature map across the entire spatial dimensions which reduces each feature map to a single value. The fourth layer, flatten, transforms the 2D feature maps into a 1D vector for further processing in dense layer. The fifth layer is a dense layer consisting of 32 neurons, and the activation function ReLU. The sixth layer is a dense layer with 1 neuron, with activation function sigmoid which outputs a probability score between 0 and 1 for binary classification. The optimization algorithm used is Adam, and the loss function is binary cross entropy for the model. Furthermore, the model with the best loss for validation is saved, the early setting is implemented with patience of 20 epochs with number of epochs set to 2000, and learning rate to Piecewise Constant Decay from $10^{-4}$ to $10^{-5}$. The number of trainable parameters are 6,177.

\FloatBarrier

\subsection{Efficient Sub-Pixel Convolutional Neural Network}
\label{ESPCNN}
\begin{figure}[H]
    \centering
    \includegraphics[width=0.9\linewidth]{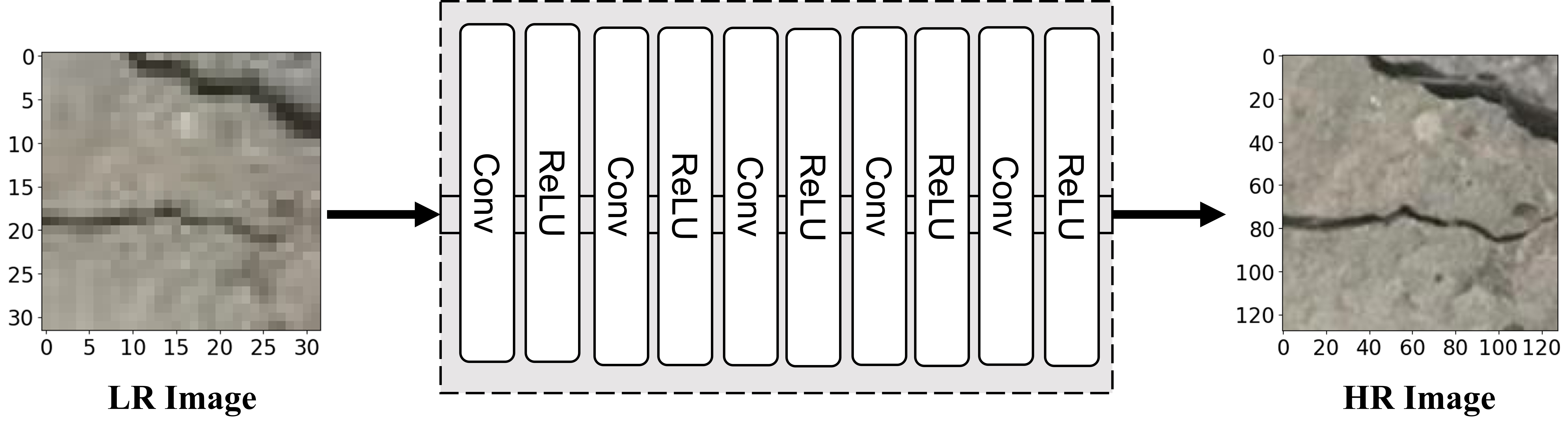}
    \caption{A schematic illustration of efficient sub-pixel convolutioal neural network (ESPCNN) architecture. The ESPCNN takes low-resolution surface image as input and generates high-resolution surface images}
    \label{fig:3}
\end{figure}

An ESPCNN is selected for its computational efficiency and competitive accuracy, as demonstrated in previous comparative studies of super-resolution techniques \cite{pawar2024esm,soltanmohammadi2023comparative}. The ESPCNN architecture as shown in Fig.~\ref{fig:3} consists of hidden convolutional layers and sub-pixel layer \cite{shi2016real}. For all convolutional layers, the SAME padding, orthogonal kernel initializer, and ReLU activation function are employed. The first convolutional layer have 64 filters, and kernel size of \(5\times5\). The second convolutional layer have kernel size of \(3\times3\), and 64 filters. The third and fourth convolutional layer have kernel size of  \(3\times3\), and 32 filters. The final layers have a $r^2 \times c$ filters, where 'c' represents the number of channels in the fine-resolution image, and 'r' signifies the upscaling ratio i,e., 4; while the kernel size is \(3\times3\). The  Adam is used as the optimization algorithm, mean squared error (MSE) is the loss function for the algorithm, and learning rate to Piecewise Constant Decay from $10^{-4}$ to $10^{-5}$. The number of trainable parameters is 83,376. Furthermore, the model with the best loss for validation is saved, early setting is implemented with patience of 20 epochs with number of epochs set to 2000.

\FloatBarrier

\section{Statistical Metrics}
For the classification, we utilize accuracy , precision , recall, and F1 score \cite{yang2021automatic,rao2021vision}. The accuracy is given by formula,

\begin{equation}
\mathrm{Accuracy} = \frac{\text{TP} + \mathrm{TN}}{\mathrm{TP} + \mathrm{TN} + \mathrm{FP} + \mathrm{FN}},
\end{equation}

\vspace{0.5cm}

where TP, TN, FP, and FN is true postive, true negative, false positive and false negative, respectively.

Precision measures the true positive predictions out of total positive predictions. It is defined as follows:

\begin{equation}
\mathrm{Precision} = \frac{\mathrm{TP}}{\mathrm{TP} + \mathrm{FP}},
\end{equation}

\vspace{0.5cm}
Recall calculates correctly predicted images out of all actual positive images, given by,

\begin{equation}
\mathrm{Recall} = \frac{\mathrm{TP}}{\mathrm{TP} + \mathrm{FN}}
\end{equation}

\vspace{0.5cm}
The F1 score is the harmonic mean of precision and recall, calculated as follows:

\begin{equation}
\mathrm{F1 score} = 2 \cdot \frac{\mathrm{Precision} \cdot \mathrm{Recall}}{\mathrm{Precision} + \mathrm{Recall}}
\end{equation}

\vspace{0.5cm}
For evaluating the super-resolution task, Absolute Point Error (APE), denoted as $E(x,y)$ \cite{pawar2024geo}, is used and defined as follows:

\begin{equation}
\mathrm{APE} = E(x,y) = |P_{1} - P_{2}|,
\end{equation}

It is the absolute difference between the reconstructed and ground truth profiles, where $P_{1}$ represents the reconstructed infrastructure image and $P_{2}$ is the ground truth image.
The other metric we utilize is peak-signal-to-noise-ratio (PSNR) given by formula,

\begin{equation}
\mathrm{PSNR} = 10 \cdot \log \frac{[max(P_1 , P_2) - min(P_1 , P_2)]^2}{MSE},
\end{equation}

where MSE is mean squared error. It is denoted in decibels (dB). PSNR quantifies proportion between highest possible power of a signal and power of corrupting noise that affects the fidelity of its representation \cite{deng2018enhancing}. Furthermore, we utilized structural-similarity index measure (SSIM), represents the contrast differences between the images, luminance, and structural similarity \cite{dosselmann2011comprehensive},

\begin{align}
\mathrm{SSIM} = \frac{(2\mu_{p_1}\mu_{p_2} + c_1)(2\sigma_{{p_1}{p_2}} + c_2)}{(\mu_{p_1}^2 + \mu_{p_2}^2 + c_1)(\sigma_{p_1}^2 + \sigma_{p_2}^2 + c_2)},
\end{align}

where $\mu_{p_1}$ and $\mu_{p_2}$ are the average of the reconstructed and ground truth images, $\sigma_{p_1}^2$ and $\sigma_{p_2}^2$  are the variances of $x_1$ and $x_2$, $\sigma_{{x_1}{x_2}}$ is the covariance between two images, and $c_1$ and $c_2$ are small constants added for numerical stability. The final metric we used is learned perceptual image patch similarity (LPIPS) \cite{zhang2018unreasonable,pawar2024esm}, defined by,

\begin{align}
\mathrm{LPIPS} = \frac{1}{M} \sum_{i=1}^M \left[ \sum_{j=1}^K w_j \left| \phi_j(x_1)_i - \phi_j(x_2)_i \right|^p \right]^\frac{1}{p},
\end{align}

here, $\phi_j$ represents feature maps from the j-th layer of a pre-trained deep neural network, $w_j$ are the applied weights for each layer, $M$ is the number of overlapping patches extracted from the profiles, $K$ is the number of layers considered, and p is 2 for Euclidean distance. The LPIPS is a more rigorous evaluation metric as it captures fine-detail dissimilarity between ground truth and generated image. This is due to it utilizing neural networks which mimic human perception of image similarity.

\section{Results}

\subsection{Training and Validation}
\begin{figure}[H]
    \centering
    \includegraphics[width=1\linewidth]{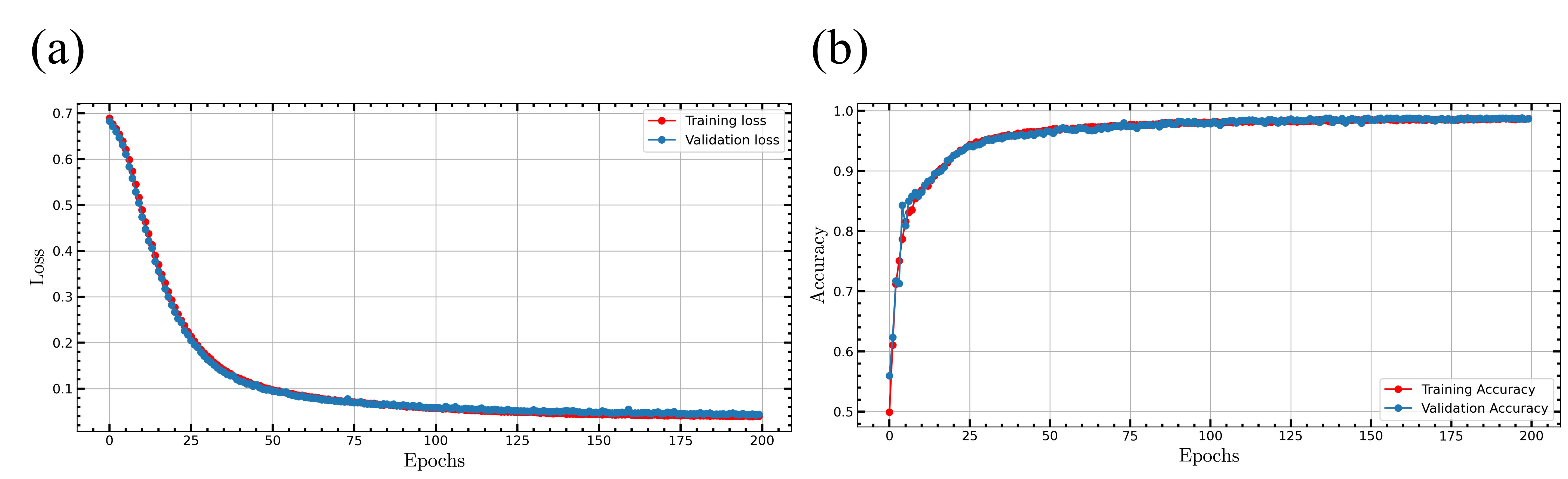}
    \caption{Training learning curves for convolution neural network. (a) The loss function plotted against the number of epochs. (b) The accuracy plotted against the number of epochs.}
    \label{fig:4}
\end{figure}

Initially, a CNN is trained to classify the low-resolution infrastructure images into two classes positive (distress), and negative (no distress). We trained the CNN with hyperparameters listed in Subsection~\ref{CNN} using training and  validation datasets. To ensure our model does not overfit or underfit we monitored the learning, i.e., loss versus epochs, and accuracy versus epochs as shown in  Fig.~\ref{fig:4}. 

\FloatBarrier

\begin{figure}[H]
    \centering
    \includegraphics[width=1\linewidth]{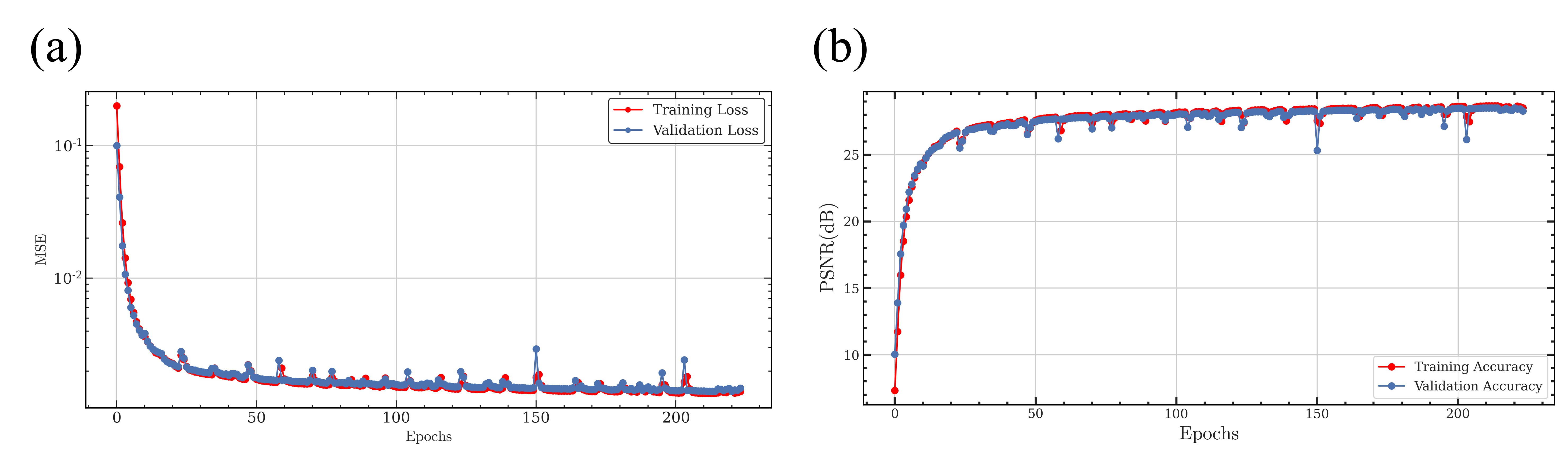}
    \caption{Training learning curves for efficient sub-pixel convolutional neural network. (a) The MSE function plotted against the number of epochs. (b) The PSNR plotted against the number of epochs}
    \label{fig:5}
\end{figure}

In the next step, we trained ESPCNN for super-resolution of positive low-resolution images to high-resolution. We train  ESPCNN with hyperparameters listed in Subsection~\ref{ESPCNN} using the training and  validation dataset. To ensure our model does not overfit or underfit we monitored the learning, i.e., mean square error (MSE) versus epochs, and peak-signal-to-noise-ratio versus epochs as shown in  Fig.~\ref{fig:5}.

\FloatBarrier

\subsection{Testing}
The trained CNN is evaluated on the test dataset i.e.,3000 belonging to both classes using various metrics such as accuracy, recall, precision , and F1 score. The statistical metrics for classification using the CNN on the test dataset are shown in Table~\ref{Table:1}. The evaluation metrics are calculated for each class and the average of each of metric is shown in Table~\ref{Table:1}. All metrics exceed 0.95, indicating an accurate classification of infrastructure into positive (indicating distress) and negative (no distress) categories.

\begin{table}[H]
   \centering
   \caption{Evaluation metrics on the test data (3000 images) for crack classification using CNN.}
   \label{Table:1}
   \begin{tabularx}{\textwidth}{>{\raggedright}p{5cm} >{\raggedright\arraybackslash}X}
     \toprule
     Metric & Value \\
     \midrule
     Accuracy & 98.667 \\
     Recall & 98.667 \\
     Precision & 98.668 \\
     F1 Score & 98.667 \\
     \bottomrule
  \end{tabularx}
\end{table}

To better understand these evaluation metrics we present confusion metrics in Table~\ref{Table:2}. The accuracy is correctly predicted in both classes which is  98.66\%. The precision, recall and F1 score is calculated for each class and average is reported. For example for positive class, the precision indicates number of correctly predicted positive infrastructure images i.e., 1497 out of all predicted positive images (1512) that equals 99.01\%. Furthermore, recall is correctly predicted positive images (1497) out of all actual positive (1522) that equals 98.36\%. F1 score is the harmonic mean of precision and recall for positive images that equals 98.69\% . For negative class, precision is 98.32\%, recall 98.99\% and F1 score is 98.65\% .

\begin{table}[H]
   \centering
   \caption{Confusion matrix generated by the CNN for the test dataset (3000 images).}
   \label{Table:2}
   \begin{tabularx}{\textwidth}{>{\raggedright}p{5cm} > {\raggedright}p{5cm} > {\raggedright\arraybackslash}X}
     \toprule
     & Predicted Positive & Predicted Negative \\
     \midrule
     Actual Positive & 1497 & \textbf{25} \\
     Actual Negative & \textbf{15} & 1463 \\
     \bottomrule
  \end{tabularx}
\end{table}

In the next step, we evaluated the trained ESPCNN for super-resolution using the test dataset (positive, i.e., indicating distress) which equals 1000 images. The evaluation metrics utilized are PSNR, SSIM and LPIPS. Furthermore, we compare the evaluation metrics of ESPCNN with bicubic interpolation as shown in Fig.~\ref{Table:3}. The results show that ESPCNN outperforms bicubic interpolation in all three metrics for super-resolution of positive LR infrastructure images. The average PSNR increase is 1.00 dB, SSIM increase is 2.32\%, and LPIPS decrease is approximately 5\% .   

\begin{table}[H]
   \centering
   \caption{Evaluation metrics on test dataset (1000 images) for super-resolution using ESPCNN.}
   \label{Table:3}
   \begin{tabularx}{\textwidth}{>{\raggedright}p{4cm} > {\raggedright}p{4cm}  > {\raggedright}p{4cm} > {\raggedright\arraybackslash}X}
     \toprule
     & PSNR & SSIM & LPIPS \\
     \midrule
     Bicubic Interpolation & 28.00 & 0.86 & 0.41 \\
     ESPCNN & 29.00 & 0.88 & 0.39 \\
     \bottomrule
  \end{tabularx}
\end{table}

\begin{figure}[H]
    \centering
    \includegraphics[width=1\linewidth]{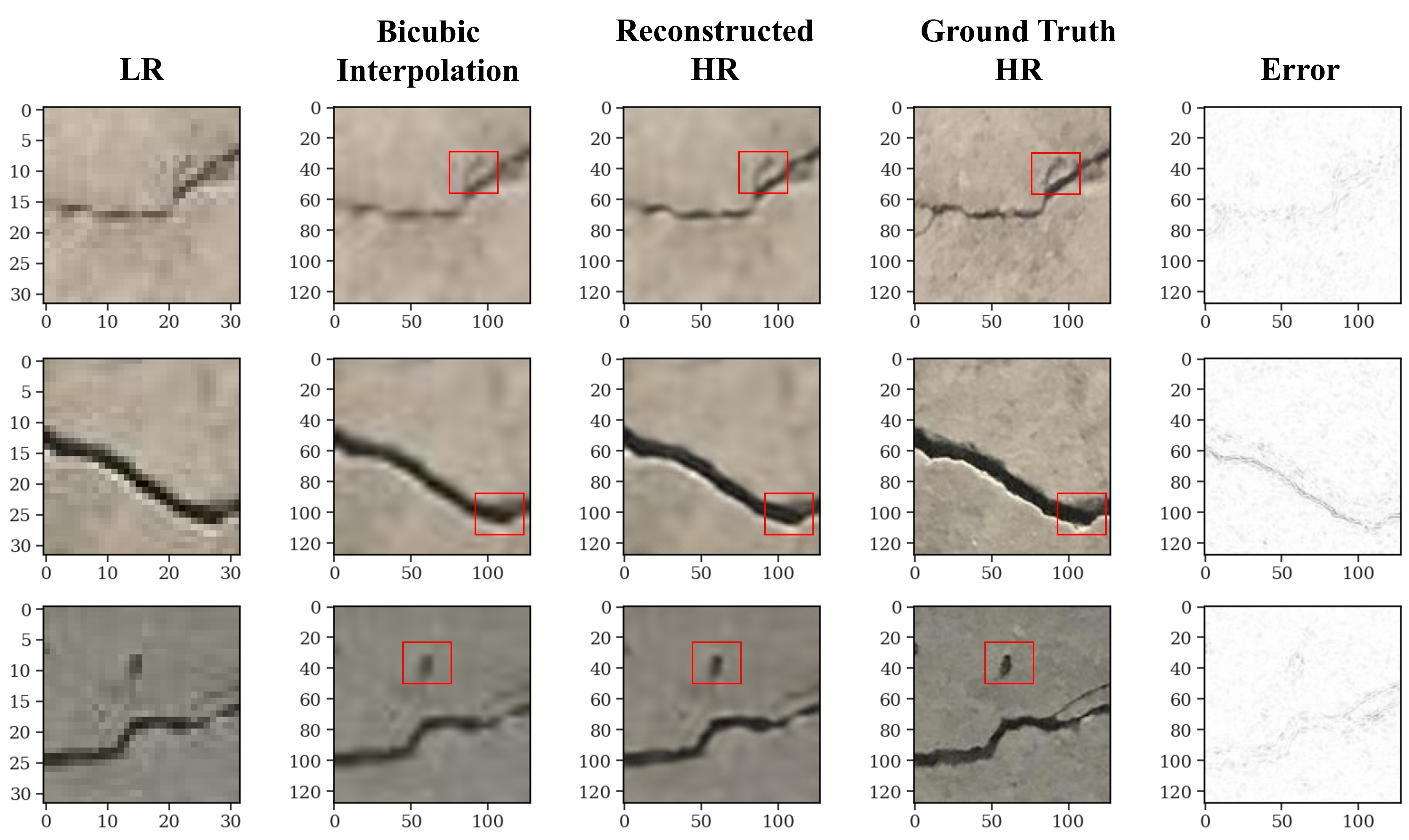}
    \caption{The comparison of HR generated using bicubic interpolation, and ESPCNN. ESPCNN demonstrates PSNR improvements compared to bicubic interpolation, from 29.08 to 30.11 dB for the first image, from 25.17 to 26.96 dB for the second image, and from 28.73 to 29.82 dB for the third image.}
    \label{fig:6}
\end{figure}

To visually evaluate the impact of using bicubic interpolation and ESPCNN for super-resolution, we plot three distinct infrastructure images from the test dataset, as shown in Fig.~\ref{fig:6}. Each image consists of the following columns: low-resolution (LR) in the first column, high-resolution (HR) generated using bicubic interpolation in the second column, HR generated using ESPCNN in the third column, ground truth HR in the fourth column, and the error between ESPCNN and ground truth HR in the fifth column. The absolute error between ESPCNN and the ground truth HR is very small, as observed in column five. In the red box of the first row, it can be seen that ESPCNN captures small distresses in HR with sharp edges and geometry compared to bicubic interpolation. In the red box of the second row, ESPCNN accurately captures crack propagation and complex geometry, whereas bicubic interpolation only captures crack propagation but fails to maintain the geometry. The red box in the third row demonstrates that ESPCNN captures even minor distresses, while bicubic interpolation creates blurs and fails to capture exact geometry.

\FloatBarrier

\section{Conclusion}
This study proposed a unique framework consisting of convolutional neural network (CNN) and efficient sub-pixel convolutional neural network (ESPCNN) to address the dual challenge of pre-processing and efficient super-resolution of infrastructure images. The CNN was designed to filter out low-resolution (LR) images containing negative cracks, while the ESPCNN was employed to superresolve LR images depicting positive distress, thereby facilitating accurate crack detection. The evaluation metric includes accuracy, precision, recall, and F1 score for the CNN, and peak-signal-to-noise ratio, structural similarity index, and learned perceptual image patch similarity (LPIPS) for the ESPCNN. The CNN and  was trained, validated and tested on positive and negative infrastructure images, total of 10000 infrastructure images (5,000 per class—positive and negative) . For test dataset (3000 images), the trained CNN achieved an accuracy and F1 score of 98.66\%, showcasing efficiency of the pre-processing negative crack images. The ESPCNN was trained, validated, and tested on 5,000 positive infrastructure images for super-resolution, demonstrating a 5\% improvement in terms of LPIPS compared to bicubic interpolation on the test set. Furthermore, after visual inspection confirmed that the ESPCNN effectively captures crack propagation, sharp edges, and complex geometries even in minor distresses compared to bicubic interpolation. The proposed framework is particularly applicable in scenarios where infrastructure images are captured using unmanned aerial vehicles, low-resolution cameras, or in areas prone to high false alarms for positive cracks.

For future studies, CNN could benefit from the incorporation of additional physical inputs, such as texture information, to enhance the classification of positive and negative infrastructure crack images. This enhancement would facilitate accurate identification in areas like rigid pavements finished with longitudinal and transverse tining. In such areas, distinguishing between tining patterns and distress using conventional visual images can be challenging. The inclusion of texture information could aid in extracting features that allow for better differentiation. Moreover, a segmentation network could be trained following super-resolution networks, to precisely identify the length and width of cracks. This approach would assist engineers in making informed decisions regarding maintenance strategies.

\section{AUTHOR CONTRIBUTIONS}
The authors confirm contribution to the paper as follows: study conception and design: Pawar, Hong, Congress, and Prozzi; data collection: Pawar ; analysis and interpretation of results: Pawar ; draft manuscript preparation: Pawar; edit and review: Hong, Congress, and Prozzi. All authors reviewed the results and approved the final version of the manuscript.

\newpage

\bibliographystyle{trb}
\bibliography{trb_template}

\begin{thebibliography}{31}
\providecommand{\natexlab}[1]{#1}

\bibitem[{{ASCE}(2021)}]{ASCE2021}
{ASCE}, \emph{America's Infrastructure Report Card 2021}, 2021.

\bibitem[{Spencer~Jr et~al.(2019)Spencer~Jr, Hoskere, and Narazaki}]{spencer2019advances}
Spencer~Jr, B.~F., V.~Hoskere, and Y.~Narazaki, Advances in computer vision-based civil infrastructure inspection and monitoring. \emph{Engineering}, Vol.~5, No.~2, 2019, pp. 199--222.

\bibitem[{Ali et~al.(2022)Ali, Chuah, Talip, Mokhtar, and Shoaib}]{ali2022structural}
Ali, R., J.~H. Chuah, M.~S.~A. Talip, N.~Mokhtar, and M.~A. Shoaib, Structural crack detection using deep convolutional neural networks. \emph{Automation in Construction}, Vol. 133, 2022, p. 103989.

\bibitem[{Ragnoli et~al.(2018)Ragnoli, De~Blasiis, and Di~Benedetto}]{ragnoli2018pavement}
Ragnoli, A., M.~R. De~Blasiis, and A.~Di~Benedetto, Pavement distress detection methods: A review. \emph{Infrastructures}, Vol.~3, No.~4, 2018, p.~58.

\bibitem[{Zakeri et~al.(2017)Zakeri, Nejad, and Fahimifar}]{zakeri2017image}
Zakeri, H., F.~M. Nejad, and A.~Fahimifar, Image based techniques for crack detection, classification and quantification in asphalt pavement: a review. \emph{Archives of Computational Methods in Engineering}, Vol.~24, 2017, pp. 935--977.

\bibitem[{Mishra et~al.(2022)Mishra, Louren{\c{c}}o, and Ramana}]{mishra2022structural}
Mishra, M., P.~B. Louren{\c{c}}o, and G.~V. Ramana, Structural health monitoring of civil engineering structures by using the internet of things: A review. \emph{Journal of Building Engineering}, Vol.~48, 2022, p. 103954.

\bibitem[{Du et~al.(2021)Du, Yuan, Xiao, and Hettiarachchi}]{du2021application}
Du, Z., J.~Yuan, F.~Xiao, and C.~Hettiarachchi, Application of image technology on pavement distress detection: A review. \emph{Measurement}, Vol. 184, 2021, p. 109900.

\bibitem[{Zhang et~al.(2017)Zhang, Wang, Li, Yang, Dai, Peng, Fei, Liu, Li, and Chen}]{zhang2017automated}
Zhang, A., K.~C. Wang, B.~Li, E.~Yang, X.~Dai, Y.~Peng, Y.~Fei, Y.~Liu, J.~Q. Li, and C.~Chen, Automated pixel-level pavement crack detection on 3D asphalt surfaces using a deep-learning network. \emph{Computer-Aided Civil and Infrastructure Engineering}, Vol.~32, No.~10, 2017, pp. 805--819.

\bibitem[{{\"O}zgenel and Sorgu{\c{c}}(2018)}]{ozgenel2018performance}
{\"O}zgenel, {\c{C}}.~F. and A.~G. Sorgu{\c{c}}, Performance comparison of pretrained convolutional neural networks on crack detection in buildings. In \emph{Isarc. proceedings of the international symposium on automation and robotics in construction}, IAARC Publications, 2018, Vol.~35, pp. 1--8.

\bibitem[{Yang et~al.(2020)Yang, Shi, Chen, and Lin}]{yang2020deep}
Yang, Q., W.~Shi, J.~Chen, and W.~Lin, Deep convolution neural network-based transfer learning method for civil infrastructure crack detection. \emph{Automation in Construction}, Vol. 116, 2020, p. 103199.

\bibitem[{Maguire et~al.(2018)Maguire, Dorafshan, and Thomas}]{maguire2018sdnet2018}
Maguire, M., S.~Dorafshan, and R.~J. Thomas, SDNET2018: A concrete crack image dataset for machine learning applications, 2018.

\bibitem[{Xu et~al.(2019)Xu, Su, Wang, Cai, Cui, and Chen}]{xu2019automatic}
Xu, H., X.~Su, Y.~Wang, H.~Cai, K.~Cui, and X.~Chen, Automatic bridge crack detection using a convolutional neural network. \emph{Applied Sciences}, Vol.~9, No.~14, 2019, p. 2867.

\bibitem[{Alexander et~al.(2022)Alexander, Hoskere, Narazaki, Maxwell, and Spencer~Jr}]{alexander2022fusion}
Alexander, Q.~G., V.~Hoskere, Y.~Narazaki, A.~Maxwell, and B.~F. Spencer~Jr, Fusion of thermal and RGB images for automated deep learning based crack detection in civil infrastructure. \emph{AI in Civil Engineering}, Vol.~1, No.~1, 2022, p.~3.

\bibitem[{Ellenberg et~al.(2016)Ellenberg, Kontsos, Moon, and Bartoli}]{ellenberg2016bridge}
Ellenberg, A., A.~Kontsos, F.~Moon, and I.~Bartoli, Bridge related damage quantification using unmanned aerial vehicle imagery. \emph{Structural Control and Health Monitoring}, Vol.~23, No.~9, 2016, pp. 1168--1179.

\bibitem[{Xiang et~al.(2022)Xiang, Wang, Deng, Shi, and Kong}]{xiang2022crack}
Xiang, C., W.~Wang, L.~Deng, P.~Shi, and X.~Kong, Crack detection algorithm for concrete structures based on super-resolution reconstruction and segmentation network. \emph{Automation in Construction}, Vol. 140, 2022, p. 104346.

\bibitem[{Pawar et~al.(2024{\natexlab{a}})Pawar, Soltanmohammadi, Mahjour, and Faroughi}]{pawar2024esm}
Pawar, N.~M., R.~Soltanmohammadi, S.~K. Mahjour, and S.~A. Faroughi, ESM data downscaling: a comparison of super-resolution deep learning models. \emph{Earth Science Informatics}, 2024{\natexlab{a}}, pp. 1--18.

\bibitem[{Pawar et~al.(2022)Pawar, Mahjour, Kalantari, and Faroughi}]{pawar2022spatiotemporal}
Pawar, N., S.~K. Mahjour, N.~K. Kalantari, and S.~Faroughi, Spatiotemporal down-scaling for multiphase flow in porous media using implicit hypernetwork neural representation. In \emph{AGU Fall Meeting Abstracts}, 2022, Vol. 2022, pp. H45M--1555.

\bibitem[{Pawar and Faroughi(2022)}]{pawar2022complex}
Pawar, N. and S.~Faroughi, Complex fluids latent space exploration towards accelerated predictive modeling. \emph{Bulletin of the American Physical Society}, Vol.~67, 2022.

\bibitem[{Bae et~al.(2021)Bae, Jang, and An}]{bae2021deep}
Bae, H., K.~Jang, and Y.-K. An, Deep super resolution crack network (SrcNet) for improving computer vision--based automated crack detectability in in situ bridges. \emph{Structural Health Monitoring}, Vol.~20, No.~4, 2021, pp. 1428--1442.

\bibitem[{Jang et~al.(2022)Jang, Jung, and An}]{jang2022automated}
Jang, K., H.~Jung, and Y.-K. An, Automated bridge crack evaluation through deep super resolution network-based hybrid image matching. \emph{Automation in Construction}, Vol. 137, 2022, p. 104229.

\bibitem[{Chen et~al.(2021)Chen, Li, Bai, Yang, Jiang, and Miao}]{chen2021review}
Chen, L., S.~Li, Q.~Bai, J.~Yang, S.~Jiang, and Y.~Miao, Review of image classification algorithms based on convolutional neural networks. \emph{Remote Sensing}, Vol.~13, No.~22, 2021, p. 4712.

\bibitem[{Soltanmohammadi and Faroughi(2023)}]{soltanmohammadi2023comparative}
Soltanmohammadi, R. and S.~A. Faroughi, A comparative analysis of super-resolution techniques for enhancing micro-CT images of carbonate rocks. \emph{Applied Computing and Geosciences}, Vol.~20, 2023, p. 100143.

\bibitem[{Zhang et~al.(2016)Zhang, Yang, Zhang, and Zhu}]{zhang2016road}
Zhang, L., F.~Yang, Y.~D. Zhang, and Y.~J. Zhu, Road crack detection using deep convolutional neural network. In \emph{2016 IEEE international conference on image processing (ICIP)}, IEEE, 2016, pp. 3708--3712.

\bibitem[{Li et~al.(2021)Li, Liu, Yang, Peng, and Zhou}]{li2021survey}
Li, Z., F.~Liu, W.~Yang, S.~Peng, and J.~Zhou, A survey of convolutional neural networks: analysis, applications, and prospects. \emph{IEEE transactions on neural networks and learning systems}, Vol.~33, No.~12, 2021, pp. 6999--7019.

\bibitem[{Shi et~al.(2016)Shi, Caballero, Husz{\'a}r, Totz, Aitken, Bishop, Rueckert, and Wang}]{shi2016real}
Shi, W., J.~Caballero, F.~Husz{\'a}r, J.~Totz, A.~P. Aitken, R.~Bishop, D.~Rueckert, and Z.~Wang, Real-time single image and video super-resolution using an efficient sub-pixel convolutional neural network. In \emph{Proceedings of the IEEE conference on computer vision and pattern recognition}, 2016, pp. 1874--1883.

\bibitem[{Yang et~al.(2021)Yang, Wang, Li, Fei, Liu, Mahboub, and Zhang}]{yang2021automatic}
Yang, G., K.~C. Wang, J.~Q. Li, Y.~Fei, Y.~Liu, K.~C. Mahboub, and A.~A. Zhang, Automatic pavement type recognition for image-based pavement condition survey using convolutional neural network. \emph{Journal of Computing in Civil Engineering}, Vol.~35, No.~1, 2021, p. 04020060.

\bibitem[{Rao et~al.(2021)Rao, Nguyen, Palaniswami, and Ngo}]{rao2021vision}
Rao, A.~S., T.~Nguyen, M.~Palaniswami, and T.~Ngo, Vision-based automated crack detection using convolutional neural networks for condition assessment of infrastructure. \emph{Structural Health Monitoring}, Vol.~20, No.~4, 2021, pp. 2124--2142.

\bibitem[{Pawar et~al.(2024{\natexlab{b}})Pawar, Soltanmohammadi, Faroughi, and Faroughi}]{pawar2024geo}
Pawar, N.~M., R.~Soltanmohammadi, S.~Faroughi, and S.~A. Faroughi, Geo-guided deep learning for spatial downscaling of solute transport in heterogeneous porous media. \emph{Computers \& Geosciences}, Vol. 188, 2024{\natexlab{b}}, p. 105599.

\bibitem[{Deng(2018)}]{deng2018enhancing}
Deng, X., Enhancing image quality via style transfer for single image super-resolution. \emph{IEEE Signal Processing Letters}, Vol.~25, No.~4, 2018, pp. 571--575.

\bibitem[{Dosselmann and Yang(2011)}]{dosselmann2011comprehensive}
Dosselmann, R. and X.~D. Yang, A comprehensive assessment of the structural similarity index. \emph{Signal, Image and Video Processing}, Vol.~5, No.~1, 2011, pp. 81--91.

\bibitem[{Zhang et~al.(2018)Zhang, Isola, Efros, Shechtman, and Wang}]{zhang2018unreasonable}
Zhang, R., P.~Isola, A.~A. Efros, E.~Shechtman, and O.~Wang, The unreasonable effectiveness of deep features as a perceptual metric. In \emph{Proceedings of the IEEE conference on computer vision and pattern recognition}, 2018, pp. 586--595.

\end{thebibliography}
\end{document}